\documentclass[runningheads]{llncs}
\usepackage{graphicx,amsmath,amsfonts,amssymb,mathtools,graphicx}
\usepackage{siunitx,arydshln,multirow}
\newcommand{\inv}{^{\raisebox{.2ex}{$\scriptscriptstyle-1$}}}

\begin{document}
\title{Semi-Supervised Medical Image Segmentation via Learning Consistency under Transformations}
\titlerunning{Semi-Supervised Segmentation via Transformation Consistency}
%
\author{
Gerda Bortsova\inst{1},
Florian Dubost\inst{1},
Laurens Hogeweg\inst{2},
Ioannis Katramados\inst{2},
Marleen de Bruijne\inst{1,3}
}


\authorrunning{G. Bortsova et al.}
%
\institute{
Biomedical Imaging Group Rotterdam, \\ Department of Radiology \& Nuclear Medicine, Erasmus MC, The Netherlands
\and 
COSMONiO, The Netherlands 
\and 
Department of Computer Science, University of Copenhagen, Denmark
}
\maketitle              
\begin{abstract}
The scarcity of labeled data often limits the application of supervised deep learning techniques for medical image segmentation.
This has motivated the development of semi-supervised techniques that learn from a mixture of labeled and unlabeled images.
In this paper, we propose a novel semi-supervised method that, in addition to supervised learning on labeled training images, learns to predict segmentations consistent under a given class of transformations on both labeled and unlabeled images.
More specifically, in this work we explore learning equivariance to elastic deformations.
We implement this through: 1) a Siamese architecture with two identical branches, each of which receives a differently transformed image, and 2) a composite loss function with a supervised segmentation loss term and an unsupervised term that encourages segmentation consistency between the predictions of the two branches.
We evaluate the method on a public dataset of chest radiographs with segmentations of anatomical structures using 5-fold cross-validation.
The proposed method reaches significantly higher segmentation accuracy compared to supervised learning.
This is due to learning transformation consistency on both labeled and unlabeled images, with the latter contributing the most.
We achieve the performance comparable to state-of-the-art chest X-ray segmentation methods while using substantially fewer labeled images.
\end{abstract}
\section{Introduction}

Supervised deep learning algorithms often require numerous labeled examples for training to yield satisfactory performance.
In the domain of medical imaging, however, labeled data is often very scarce, which is especially true for dense labels such as segmentations, as they are particularly costly to produce.
On the other hand, for many medical image analysis tasks, unlabeled data coming from the same or similar distribution as the labeled data is available in abundance.
This motivates the development of semi-supervised algorithms.

Many ideas have been proposed for improving performance of deep learning algorithms for medical image analysis through utilizing unlabeled data \cite{cheplygina2019not}.
Self-supervised approaches learn to perform an auxiliary task related to the target prediction task on unlabeled data, examples including image colorization \cite{ross2018endoscopy} and predicting one image modality from another \cite{hervella2018retinal}.
Auxiliary manifold embedding \cite{baur2017semi} learns to place unlabeled examples far or close in the feature space with respect to each other and to labeled examples according to prior adjacency information.
Self-ensembling approaches learn from synthetic labels on unlabeled data, which are constructed using past iterations of the same network \cite{Li2018,Perone2018}.

In this paper, we propose an approach in which we learn consistency under transformations on both labeled and unlabeled data, in addition to supervised learning from labeled data.
We implement this consistency learning through a Siamese architecture trained end-to-end.
The network has two identical branches each of which receives differently transformed versions of the same images as inputs and is supervised to output segmentations consistent with the other branch, in addition to learning from labeled images in a supervised fashion.
Combining supervised and unsupervised consistency learning in the network is achieved by a composite objective consisting of two respective terms.
This idea was already explored before and applied to the prediction of image-level labels: image classification \cite{Sajjadi2016,Rasmus2015} and landmark coordinate regression \cite{Honari2018}.
We extend this method so that it can be applied efficiently to learning to predict pixel-level labels consistent under spatial transformations of images.
This entails including a special differentiable layer into the Siamese architecture that transforms the pixel-wise predictions of one of the branches so as to align it with the predictions of the second branch and consequently allow their pixel-wise comparison in the consistency loss term.
Self-ensembling approaches \cite{Li2018,Perone2018} are similar to the proposed approach in that they use a transformation consistency prior: they construct their synthetic labels on unlabeled data using predictions on differently transformed inputs.
Unlike these methods, the proposed Siamese approach does not train the network to fit specific targets on unlabeled images (i.e. the synthetic labels), which are unknown and cannot be reliably estimated; it only encourages the outputs to have the desired transformation consistency property.

We evaluate our method on the JSRT chest X-ray dataset \cite{jsrt,Bram2006}.
In this paper, we focus on learning equivariance to elastic deformations, although our method can be readily applied to a broader class of transformations.
Through our experiments, we evaluate: 1) the contribution of learning this equivariance on labeled data (i.e. as a regularization in supervised-only learning) to the segmentation performance; 2) the contribution of adding different amounts of unlabeled data into the equivariance learning; 3) how these contributions vary with the size of the labeled portion of the training set.
We compare the proposed method trained in the small data (20 labeled images) and full supervision regimes with state-of-the-art methods \cite{shah2018ms,dai2018scan,novikov2018,frid2018} and the inter-observer agreement \cite{Bram2006}.

\section{Method}

Let $\mathcal{X}_l$ be a set of training examples with corresponding ground truth labels $\mathcal{Y}$ and $\mathcal{X}_u$ be a set of unlabeled examples.
Let $\mathcal{T}$ be a distribution of tuples of mappings such that for a tuple $(t^{in}, t^{out}) \eqqcolon  t \sim \mathcal{T}$ applying transformation $t^{in}$ to any image $x \in \mathcal{X} \coloneqq \mathcal{X}_u \cup \mathcal{X}_l$ would result in the corresponding label $y$ being transformed into $t^{out}(y)$ and $t^{out}$ is invertible.
Let $\mathcal{X}^\mathcal{T}_l$ be a set of all images from $\mathcal{X}_l$ with corresponding labels, augmented by examples ($t^{in}(x)$, $t^{out}(y)$).
We would like to find parameters $\theta$ of a network $f$ that optimize the following objective:
\[
\min{
\mathcal{L}_{sup}^\mathcal{T}(\theta) + \lambda \mathcal{L}_{cons}^\mathcal{T}(\theta)
}
\]
with $\mathcal{L}_{sup}^\mathcal{T}(\theta) = 1 / |\mathcal{X}^{\mathcal{T}}_l| ( \sum_{(x, y) \in \mathcal{X}^{\mathcal{T}}_l}{\mathcal{S}(y, f(x; \theta))} )$ being a regular supervised loss (using $\mathcal{T}$ as a data augmentation strategy and $\mathcal{S}$ as an image-wise loss) and $\mathcal{L}_{cons}^\mathcal{T}(\theta)$ being an unsupervised consistency loss defined as: 
\[
\mathcal{L}_{cons}^\mathcal{T}(\theta) = 
\frac{1}{|\mathcal{X}|}
\sum_{x \in \mathcal{X}}{\mathbb{E}_{t_1, t_2 \sim \mathcal{T}}
\Big[
\mathcal{C} \Big( (t^{out}_2 \circ 
{(t^{out}_1)}\inv)(f(t^{in}_1(x); \theta)), f(t^{in}_2(x); \theta) \Big)
\Big]
}
\]
$\mathcal{L}_{cons}^\mathcal{T}(\theta)$ encourages the selection of $\theta$ that maximizes consistency of network predictions under transformations $\mathcal{T}$ on $\mathcal{X}$ as measured by image-wise loss $\mathcal{C}$.

\begin{figure}[t!]
\centering
\includegraphics[width=1.0\textwidth]{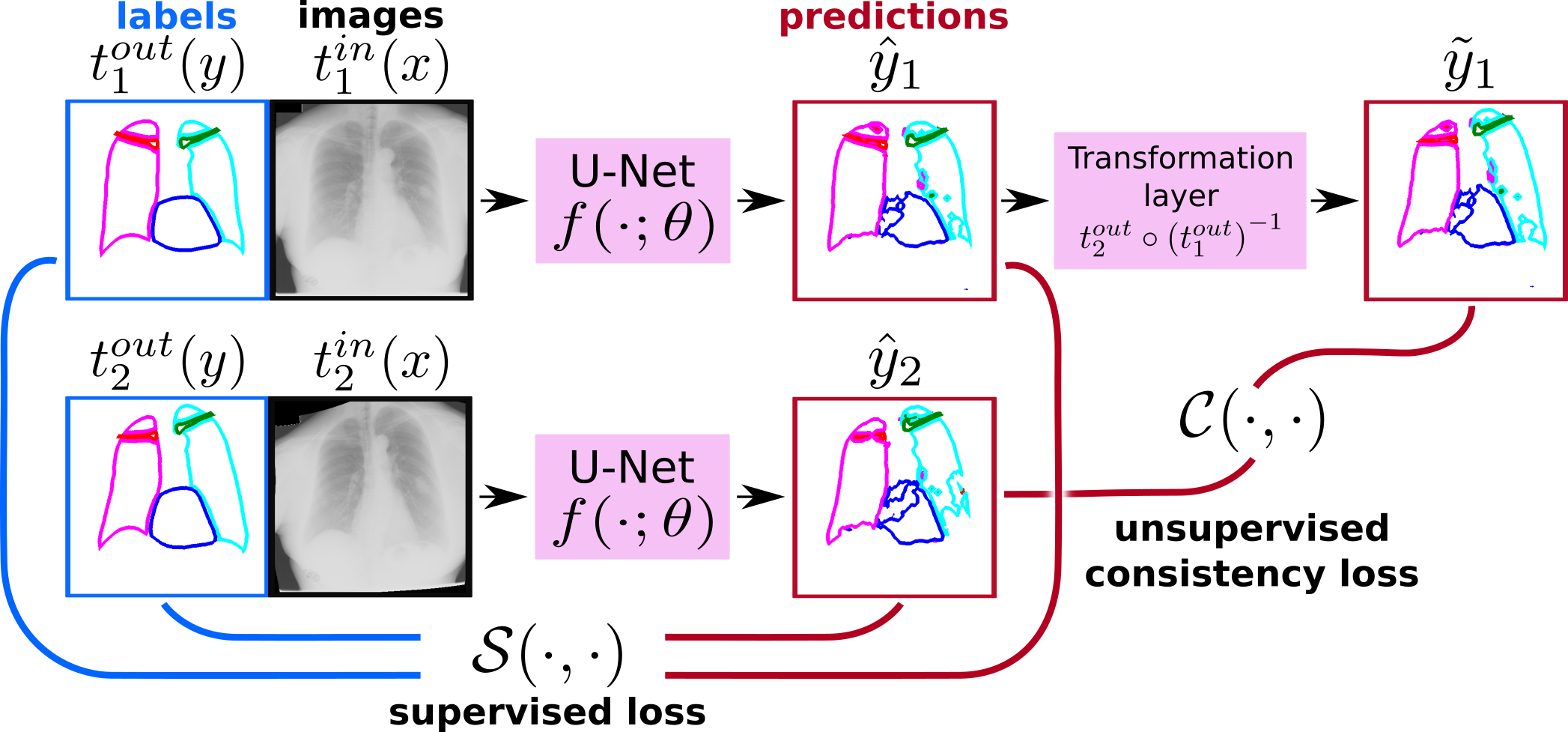}
\caption{
The proposed network.
The inputs are mixed batches of labeled and unlabeled images.
Every image $x$ is transformed by two random mappings $t^{in}_1$ and $t^{in}_2$.
The label $y$, if available, is transformed by $t^{out}_1$ and $t^{out}_2$.
$t^{in}_1(x)$ and $t^{in}_2(x)$ are fed to the two identical branches of the network.
The output of one of the branches is transformed by a differentiable layer for comparison with the output of the second branch in the consistency loss $\mathcal{C}$.
The network is trained end-to-end using a combination of $\mathcal{C}$ and a supervised loss $\mathcal{S}$ (defined only on labeled images) as specified by Eq. 1.
} \label{deformnet}
\end{figure}

We approximate minimization of this objective by a mini-batch training scheme in which we sample a set of labeled examples $\mathcal{B}_l$, a set of unlabeled examples $\mathcal{B}_u$ and two transforms $t_1, t_2 \sim \mathcal{T}$ for every example.
A member of a batch $\mathcal{B}$ thus is a tuple $(x, t_1, t_2)$ with or without ground truth label $y$.
The mini-batch objective is:
\begin{multline}
\!\!\!\!
\mathcal{L}^{\mathcal{T}\!, \mathcal{B}}(\theta) =
\frac{1}{2|\mathcal{B}_l|}
\sum_{(x, y, t_1, t_2) \in \mathcal{B}_l}{
\!\!\!\!
\Big( \mathcal{S} \big( t^{out}_1(y), f(t^{in}_1(x); \theta) \big) + \mathcal{S} \big( t^{out}_2(y), f(t^{in}_2(x); \theta) \big) \Big) } \: + \\
\frac{\lambda}{|\mathcal{B}|}
\sum_{(x, t_1, t_2) \in \mathcal{B}}{\mathcal{C} \big( (t^{out}_2 \circ (t^{out}_1)\inv)(f(t^{in}_1(x); \theta)), f(t^{in}_2(x); \theta) \big) }
\end{multline}
\label{batchLoss}%
The first sum approximates $\mathcal{L}_{sup}^\mathcal{T}(\theta)$ and the second approximates $\mathcal{L}_{cons}^\mathcal{T}(\theta)$.
In the rest of the paper we will abuse notations $\mathcal{L}_{sup}^\mathcal{T}$ and $\mathcal{L}_{sup}^\mathcal{T}$ to refer to the first and the second sums in Eq. 1, respectively.

The overview of the network implementing this scheme is shown in Fig. \ref{deformnet}.
The architecture has two branches with shared weights $\theta$.
For every example $(x, t_1, t_2) \in \mathcal{B}$, we feed two differently deformed versions of $x$ ($t^{in}_1(x)$ and $t^{in}_2(x)$) to the two branches of the network.
The output of the first branch, $\hat{y}_1 \coloneqq f(t^{in}_1(x);\theta)$, is transformed by $t^{out}_2 \circ {(t^{out}_1)}\inv$ using a custom differentiable layer so as to align it with the prediction of the second branch $\hat{y}_2 \coloneqq f(t^{in}_2(x);\theta)$ for pixel-wise comparison by the consistency term $\mathcal{C}$.
In addition to $\mathcal{C}$, if $x$ happens to be labeled, the supervised loss $\mathcal{S}$ is applied to both $\hat{y}_1$ and $\hat{y}_2$.
The network is thus trained end-to-end using Eq. 1 as a composite loss.

Note that since our transformation layer is differentiable, the gradient can flow through both branches.
If the layer did not let the gradient through, training the network would be equivalent to applying the network to $t^{in}_1(x)$ and using  $\tilde{y}_1 \coloneqq (t^{out}_2 \circ {(t^{out}_1)}\inv)(\hat{y}_1)$ as a target for $t^{in}_2(x)$ (such approach was adopted by \cite{Li2018}).
In this case, the network is forced to update the prediction for $t^{in}_2(x)$ to be more similar to $\tilde{y}_1$, even if $\tilde{y}_1$ is incorrect.
In the case of a differentiable layer, the network has the freedom to update its predictions in any way that optimizes $\mathcal{C}$, which includes changing the prediction for $t^{in}_1(x)$ to be more consistent with that for $t^{in}_2(x)$, the other way around or changing both of them in the same direction.
In other words, the proposed methodology encourages predictions to have the desired property of transformation consistency without encouraging any specific predictions for unlabeled images, which might otherwise introduce a bias when the targets for these images cannot be reliably inferred.

In this work, we use elastic deformations as the event space for $\mathcal{T}$.
Our transformation layer, in addition to the predicted segmentation, takes as its inputs deformation fields specifying forward and backward transformations $t^{out}_2 \circ {(t^{out}_1)}\inv$ and $t^{out}_1 \circ {(t^{out}_2)}\inv$.
The latter is necessary for backpropagating the gradients through the layer:
\[
\frac{\partial \mathcal{C}(\hat{y}_1, \hat{y}_2)}{\partial \hat{y}_1}) = (t^{out}_1 \circ {(t^{out}_2)}\inv)
\Big(
\frac{\partial \mathcal{C}(\tilde{y}_1, \hat{y}_2)}{\partial \tilde{y}_1}
\Big)
\]
In principle, any transformations $t^{in}$ and $t^{out}$ could be implemented, as long as the inverse ${(t^{out})}\inv$ can be computed.

\section{Experiments}

\subsubsection{Dataset and validation}
We used the Japanese Society of Radiological Technology (JSRT) dataset \cite{jsrt}, which contains 247 posterior-anterior chest radiographs with a resolution $2048 \times 2048$, \SI{0.175}{\milli\meter} pixel size and 12-bit depth.
Segmentations of the left and right lung fields, left and right clavicles and the heart were made available by \cite{Bram2006}.
We created five splits of the dataset by choosing images with either even or odd IDs as the test set ($|\mathcal{X}^{\text{test}}| = 123$ or 124) as prescribed by \cite{Bram2006} and randomly splitting the rest into the training ($|\mathcal{X}^{\text{train}}| = 100$) and validation portions.
For every split, we sampled subsets $\mathcal{X}^{\text{train}}_l$ of 5, 10, 25 and 50 examples (with larger subsets containing smaller ones) from $\mathcal{X}^{\text{train}}$  to be used as labeled portions of the training set.
The rest of the images from $\mathcal{X}^{\text{train}}$ were assigned to the unlabeled portion of the training set $\mathcal{X}^{\text{train}}_u$, to be used by the proposed semi-supervised algorithm.

\subsubsection{Implementation details}
We used a U-Net-like architecture \cite{unet} as the basis for the Siamese network.
For every cross-validation (CV) split and labeled-unlabeled training set split, we first trained a network using $\mathcal{L}_{sup}^{\mathcal{T}}$ loss (i.e. in a purely supervised way).
We used this network as a basis for fine-tuning four different networks: two supervised and two semi-supervised.
The two supervised networks were fine-tuned using $\mathcal{L}_{sup}^{\mathcal{T}}$ or $\mathcal{L}_{sup}^{\mathcal{T}}+\lambda\mathcal{L}_{cons}^{\mathcal{T}}$.
We refer to them as ``the baseline'' and the supervised transformation-consistent network or \textit{SupTC}, respectively.
The semi-supervised networks were fine-tuned using $\mathcal{L}_{sup}^{\mathcal{T}}+\lambda\mathcal{L}_{cons}^{\mathcal{T}}$, with batches containing equal numbers of labeled and unlabeled examples. (The total batch size was the same as in the supervised cases.)
One of the semi-supervised networks only used unlabeled images from the training set $\mathcal{X}^{\text{\text{train}}}_u$ (we dubbed it \textit{SemiTC}), while the other one additionally used images from the corresponding validation and test sets as unlabeled (\textit{SemiTC+}).
We used intersection over union (IOU) averaged over six classes (the five structures and the background) as both supervised and unsupervised loss terms: $\mathcal{S}(y, \hat{y}) = \mathcal{C}(y, \hat{y}) = 1 / 6 \sum_c ( \sum_i{y_c^{(i)}\hat{y}_c^{(i)}}
/
(\sum_i{y_c^{(i)} + (1 - y_c^{(i)})\hat{y}_c^{(i)}}))
$.
The weight $\lambda$ of the consistency term was arbitrarily set to 1, giving the supervised and consistency terms equal importance.
Adadelta optimizer was used for both training and fine-tuning.
The images and segmentation maps were subsampled to a resolution of $512 \times 512$ for training.
The deformation fields for elastic deformations were created by randomly sampling two-dimensional displacement maps from a uniform distribution $\mathcal{U}(-1000, 1000)$ and smoothing them with a Gaussian filter with the standard deviation of 100 pixels.
Spline interpolation was applied to images and nearest neighbor interpolation was applied to labels and predictions.
To reduce computational time, probability distribution $\mathcal{T}$ was specified as drawing an identity transform $(\text{id}_\mathcal{X}, \text{id}_{\mathcal{Y} \cup \mathcal{\hat{Y}}})$ or a random elastic deformation $(t^{in}, t^{out})$ (specified by a deformation field sampled as described above) with 50\% chance.
The transformation layer was implemented in Tensorflow, which allows implementing operations with custom gradients.
Gradient backpropagation through the layer was implemented by copying gradients with respect to the layer output pixels to the positions where those pixels' values came from in the forward pass, which is elastic deformation with nearest neighbor interpolation.
Pixel values that are not copied in the forward pass receive no gradient.

\section{Results and Discussion}

Table \ref{semisup} compares the four versions of the proposed network. The metric used is IOU averaged over the five anatomical structures (mIOU).

The consistency term $\mathcal{L}_{cons}^\mathcal{T}$ improved the performance even when all training images were labeled.
Although this improvement was modest, it was very reliable: $\mathcal{L}_{cons}^\mathcal{T}$ improved mIOU in 24 experiments out of 25 (5 CV splits $\times$ 5 training set sizes).
Interestingly, \textit{SupTC} networks achieved similar or higher supervised training loss $\mathcal{L}_{sup}^{\mathcal{T}}$ most of the time, while expectedly having lower consistency loss $\mathcal{L}_{cons}^{\mathcal{T}}$ and lower total loss $\mathcal{L}_{sup}^{\mathcal{T}}+\mathcal{L}_{cons}^{\mathcal{T}}$, compared to their non-consistency-regularized counterparts.
This rules out a hypothesis that the consistency term merely helps the network to converge to a lower $\mathcal{L}_{sup}^\mathcal{T}$ (e.g. because of an increase in the learning rate).
We believe that this performance gain can be explained by that image-to-segmentation mappings that are more consistent under elastic deformations are more likely to be correct even if the resulting segmentations of training images fit less to the ground truth (which might be wrong or ill-defined).

\begin{table}[!t]
\caption{
Means and standard deviations of mIOU over the five test sets corresponding to different versions of the method (rows) and labeled training set sizes (columns).
These versions are: the proposed architecture trained with $\mathcal{L}_{sup}^{\mathcal{T}}$ only, the consistency-regularized version of the latter (\textit{SupTC}), and the proposed semi-supervised method using either only unlabeled examples from the training set (\textit{SemiTC}) or additionally validation and test set as unlabeled examples (\textit{SemiTC+}).
Note that with the largest training set size \textit{SupTC} is equivalent to \textit{SemiTC}, since all labels are available.
}
\label{semisup}
\begin{tabular}{|c|c|c|c|c|c|c|c|}
\hline
Methods & Loss & $\mathcal{X}_u$ &	5 &	10 &	25 &	50 & 100 \\
\hline
Baseline & $\mathcal{L}_{sup}^{\mathcal{T}}$ & $\varnothing$ &	$74.2 \pm 3.8$ &	82.8 $\pm$ 1.3 &	87.5 $\pm$  0.4 &	89.0 $\pm$  0.3 &	90.6 $\pm$ 0.2 \\
\hline
\textit{SupTC} &
\parbox[c]{7mm}{\multirow{3}{*}{\rotatebox[origin=c]{90}{
\scalebox{1}{
$\substack{\mathcal{L}_{sup}^{\mathcal{T}} + \;\;\;\; \\
\;\;\;\;\;\;\; \mathcal{L}_{cons}^{\mathcal{T}}}$
}}}}
 & $\varnothing$  &	76.4 $\pm$ 3.8 &	83.6 $\pm$ 1.4 &	87.8 $\pm$ 0.4 &	89.5 $\pm$ 0.2 &	90.9 $\pm$ 0.3 \\
\cline{1-1}\cline{3-8}
\textit{SemiTC} &  & $\mathcal{X}^{\scalebox{.7}[0.7]{train}}_u$ & 	85.4 $\pm$ 1.0 &	86.9 $\pm$ 1.4 &	88.7 $\pm$ 1.0 &	89.7 $\pm$ 0.2 & - 	\\
\textit{SemiTC+} &  &
\scalebox{.75}[.8]{
$\!\!\mathcal{X}^{\scalebox{0.9}[1]{train}}_u\!\!$ \scalebox{.7}[1.1]{$\cup$} $\!\!\mathcal{X}^{\scalebox{0.9}[1]{val+test}}\!\!$
}
&	85.0 $\pm$ 2.8 &	87.9 $\pm$ 0.8 &	89.7 $\pm$ 0.4 &	90.5 $\pm$ 0.3 &	91.1 $\pm$ 0.1 \\
\hline
\end{tabular}
\end{table}

The proposed semi-supervised approach \textit{SemiTC} outperformed the supervised \textit{SupTC} substantially when the size of the labeled training set was small (5 or 10 images).
This improvement was also very consistent: \textit{SemiTC} was better than \textit{SupTC} in all five CV splits.
For larger training set sizes, the improvement was more modest but still consistent (at least 4 out of 5 CV splits).
With \textit{SemiTC+}, which added validation and test images to the pool of unlabeled images for training, we achieved an additional small but consistent improvement (in 20 experiments out of 25).
The comparison of the performance gains achieved by adding $\mathcal{L}_{cons}^{\mathcal{T}}$ to the loss (i.e. the improvement of \textit{SupTC} over the baseline) and introducing unlabeled images to the training (i.e. the improvement of \textit{SemiTC} and \textit{SemiTC+} over \textit{SupTC}) suggests that the latter is mainly responsible for the superior performance of the proposed method compared to the baseline.

The proposed method substantially outperformed MS-Net \cite{shah2018ms}, the only weakly supervised method evaluated on JSRT that is known to us.
MS-Net achieved 67\% and 81\% mIOU (extracted from Fig. 4 in \cite{shah2018ms}) when trained in 20\% and 100\% strong supervision (124(123) labeled training images) modes, respectively.
(In the former case, bounding boxes and landmarks were used as labels for the remaining 80\% of the images.)
\textit{SemiTC} reached $87 \pm 1.5$\% mIOU in $<$20\% supervision mode (10 labeled images for training and 10 for validation).

Table \ref{stateofart} compares our baseline network and the proposed  \textit{SemiTC+} with the inter-observer agreement \cite{Bram2006} and state-of-the-art chest X-ray segmentation methods \cite{dai2018scan,novikov2018,frid2018}.
All these methods are based on fully convolutional networks and are trained in a supervised way using at least 124(123) labeled images from the JSRT dataset. For these comparisons, we post-processed all predicted segmentations as described in \cite{frid2018} (small objects removal, hole filling).

Both our baseline and \textit{SemiTC+} trained using 124(123) labeled images outperformed Dai et al. \cite{dai2018scan} and Novikov et al. \cite{novikov2018} in segmentation of all structures (without post-processing as well).
Both methods performed similarly to the method of Frid-Adar et al. \cite{frid2018}, with the heart segmentation being slightly worse and clavicle segmentation being slightly better.
(Note that the network of Frid-Adar et al. \cite{frid2018} benefited from pre-training on ImageNet.)
We reached human-level performance in lung and heart segmentation and approached it closely in clavicle segmentation, unlike all other methods, which had a larger gap between their performance and the observers' for clavicle segmentation.

\begin{table}[!t]
\caption{
The comparison of the inter-observer agreement, state-of-the-art techniques, our baseline and \textit{SemiTC+} trained on 20 or 124(123) labeled images.
The metrics reported are means and standard deviations of per-structure IOU and mean absolute contour distance (MACD) averaged over CV splits.
Dai et al. \cite{dai2018scan} and Novikov et al. \cite{novikov2018} did not report MACD.
}
\label{stateofart}
\setlength{\tabcolsep}{4pt}
\centering
\begin{tabular}{c|c|c|c|c|c|c}
 & Methods & $|\mathcal{X}^{\scalebox{.7}[0.7]{train+val}}_l|$ & $|\mathcal{X}^{\scalebox{.7}[0.7]{train}}_u|$ & Lungs & Heart & Clavicles \\
\hline\hline
\parbox[t]{2mm}{\multirow{8}{*}{\rotatebox[origin=c]{90}{\bfseries IOU, \%}}} & Human \cite{Bram2006} &  &  & 94.6$\pm$1.8 & 87.8$\pm$5.4 & 89.6$\pm$3.7 \\
\cline{2-7}
 & Dai et al. \cite{dai2018scan} & 209 &  & 94.7$\pm$0.4 & 86.6$\pm$1.2 &   \\
 & Novikov et al. \cite{novikov2018} & 165 &  & 94.8  & 87.8  & \textbf{85.9}  \\
 & Frid-Adar et al. \cite{frid2018} & 124(123) &  &  \textbf{96.1$\pm$1.4}  &  \textbf{90.6$\pm$3.8}  &  85.5$\pm$4.5  \\
\cline{2-7}
 & Supervised baseline & 20 &  & 93.3$\pm$3.4 & 81.7$\pm$13.1 & 75.7$\pm$12.4 \\
 & \textit{SemiTC+} & 20 & 237 & 94.5$\pm$2.1 & 85.2$\pm$9.4 & 83.3$\pm$6.6 \\
 & Supervised baseline & 124(123) &  & 95.3$\pm$2.3 & \textbf{88.8$\pm$5.2} & 87.3$\pm$4.6 \\
 & \textit{SemiTC+} & 124(123) & 147 &  \textbf{95.5$\pm$1.9}  &  \textbf{88.8$\pm$4.9}  &  \textbf{88.1$\pm$4.4}  \\
\hline\hline
\parbox[t]{2mm}{\multirow{6}{*}{\rotatebox[origin=c]{90}{\bfseries MACD, \SI{}{\milli\meter}}}} & Human \cite{Bram2006} &  &  & 1.64$\pm$0.69 & 3.78$\pm$1.82 & 0.68$\pm$0.26 \\
\cline{2-7}
 & Frid-Adar et al. \cite{frid2018} & 124(123) &  &  \textbf{1.02$\pm$0.56}  &  \textbf{2.54$\pm$1.13}  &  \textbf{0.85$\pm$0.32}  \\
\cline{2-7}
 & Supervised baseline & 20 &  & 1.67$\pm$0.99 & 5.56$\pm$4.41 & 1.93$\pm$2.42 \\
 & \textit{SemiTC+} & 20 & 237 & 1.35$\pm$0.58 & 4.51$\pm$3.20 & 1.16$\pm$0.57 \\
 & Supervised baseline & 124(123) &  & 1.17$\pm$0.70 & \textbf{3.36$\pm$1.61} & 0.87$\pm$0.39 \\
 & \textit{SemiTC+} & 124(123) & 147 &  \textbf{1.10$\pm$0.57}  &  \textbf{3.37$\pm$1.58}  &  \textbf{0.81$\pm$0.32}  \\
\end{tabular}
\end{table}

\textit{SemiTC+} trained only on 20 images (10 for training and 10 for validation) reached human-level performance in lung segmentation and was only slightly worse than the observers in heart segmentation (2.6\% lower IOU).
Its clavicle segmentation performance was substantially worse than human, but was only slightly worse than the automatic methods \cite{novikov2018,frid2018} trained using the fully labeled dataset  (2.6\% and 2.2\% lower IOU, respectively).
This could not be achieved by purely supervised training with the small labeled set, which was substantially worse in segmentation of all structures.

\section{Conclusion}

We proposed a novel semi-supervised segmentation method that learns consistency under transformations.
The evaluation on a public chest X-ray dataset showed that the proposed consistency regularization improved the segmentation performance both when all training data was labeled and when additional unlabeled data was used for training.
We achieved the performance comparable to the state-of-the-art while using more than five times fewer labeled images. 

\subsubsection{Acknowledgments}
This research is part of the research project Deep Learning for Medical Image Analysis (DLMedIA) with project number P15-26, funded by 
the Netherlands Organisation for Scientific Research (NWO).
The computations were carried out on the Dutch national e-infrastructure with the support of SURF Cooperative.

%
%

\bibliographystyle{splncs04}
\bibliography{bib}

\end{document}